\pdfoutput=1
\documentclass[letterpaper, 10 pt, conference]{IEEEtran}

\IEEEoverridecommandlockouts

\usepackage{graphicx}
\usepackage{amsmath}
\usepackage{subcaption}    
\usepackage{stfloats}
\usepackage{booktabs} 
\usepackage{threeparttable} 
\usepackage{multirow} 
\usepackage[letterpaper, left=0.75in, right=0.75in, top=0.75in, bottom=0.75in]{geometry}

\usepackage{cite}

\usepackage{eso-pic}

\newcommand{\addIEEEcopyrightnotice}{%
  \AddToShipoutPictureFG*{%
    \AtPageUpperLeft{%
      \raisebox{-0.34in}{%
        \hspace{0.75in}%
        \makebox[0.92\paperwidth][l]{\fontsize{6.7}{7.5}\selectfont 2026 IEEE INTERNATIONAL CONFERENCE ON ROBOTICS AND AUTOMATION (ICRA). PREPRINT VERSION. ACCEPTED FOR ICRA 2026.}%
      }%
    }%
    \AtPageLowerLeft{%
      \raisebox{0.12in}{%
        \makebox[\paperwidth][c]{%
          \parbox{0.92\paperwidth}{%
            \centering\fontsize{5.5}{6.5}\selectfont
            \copyright{} 2026 IEEE. Personal use of this material is permitted.
            Permission from IEEE must be obtained for all other uses, in any current or future media,
            including reprinting/republishing this material for advertising or promotional purposes,
            creating new collective works, for resale or redistribution to servers or lists,
            or reuse of any copyrighted component of this work in other works.%
          }%
        }%
      }%
    }%
  }%
}

\title{\vspace{0.25in} The Curse of Precision: A Data Scaling Law for High-Precision Robotic Manipulation
}

\author{Cuijie Xu$^1$, Yuanfan Xu$^1$, Min Xue$^1$, Jianjie Lin$^3$, Jian Wang$^1$, Xudong Zhang$^1$, Yu Wang$^1$, Jincheng Yu$^{1,2,*}$
\thanks{$^*$ \textbf{Corresponding Author.}}
\thanks{$^1$Department of Electronic Engineering, Tsinghua University, Beijing, China. xcj20@mails.tsinghua.edu.cn, yu-jc@mail.tsinghua.edu.cn}
\thanks{$^2$Institute for Embodied Intelligence and Robotics, Tsinghua University.}
\thanks{$^3$OpenMind (WuHu) Robotics Co., Ltd., Wuhu, China.}
\thanks{This research was supported by National Natural Science Foundation of China (62325405), Tsinghua University Initiative Scientific Research Program, Tsinghua-Efort Joint Research Center for EAI Computation and Perception and SunRisingAI Lab, Beijing National Research Center for Information Science, Technology (BNRist), Beijing Innovation Center for Future Chips, and State Key laboratory of Space Network and Communications.}
}
\begin{document}

\addIEEEcopyrightnotice
\maketitle
\thispagestyle{empty}
\pagestyle{empty}

\begin{abstract}

While scaling laws for imitation learning have primarily focused on generalization in open-world settings, the relationship between data and precision in closed-world tasks like robotic assembly remains largely unexplored. This paper systematically investigates this relationship and introduces a novel scaling law. We find that to achieve a fixed success rate, the required number of demonstrations $N$, grows super-exponentially as the target precision $P$, approaches a limit $c$. This relationship is accurately captured by the model $\log(N) \propto 1/(P-c)$. Crucially, we reveal that the limit precision $c$ is not a static physical constant of the task but an emergent property of the entire agent system, including its sensors and expert policy. Through experiments on canonical manipulation tasks, we validate this law and demonstrate that improving system components—such as adding a wrist camera or using a more effective expert—measurably lowers $c$, thus expanding the system's achievable precision. Our work provides a new theoretical framework for precision in robotics and a quantitative metric to evaluate system capabilities. Furthermore, these findings provide a practical methodology for guiding the development and debugging of high-precision manipulation systems.

\end{abstract}

\section{Introduction}

\begin{figure*}[t]
    \centering 

    \begin{subfigure}{0.3\textwidth} 
        \centering
        \includegraphics[width=0.95\linewidth]{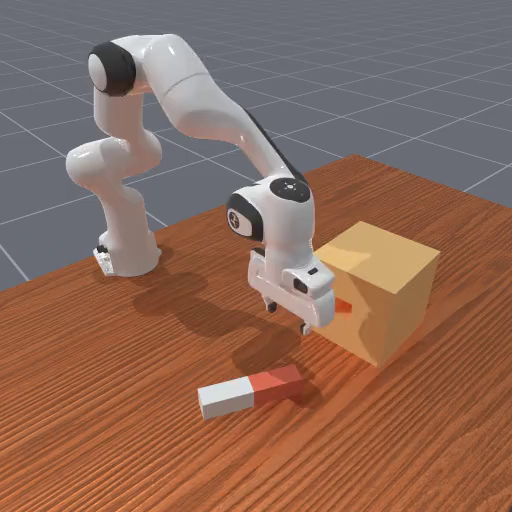} 
        \caption{Peg Insertion}
        \label{fig:task_peg_insertion}
    \end{subfigure}
    \begin{subfigure}{0.3\textwidth}
        \centering
        \includegraphics[width=0.95\linewidth]{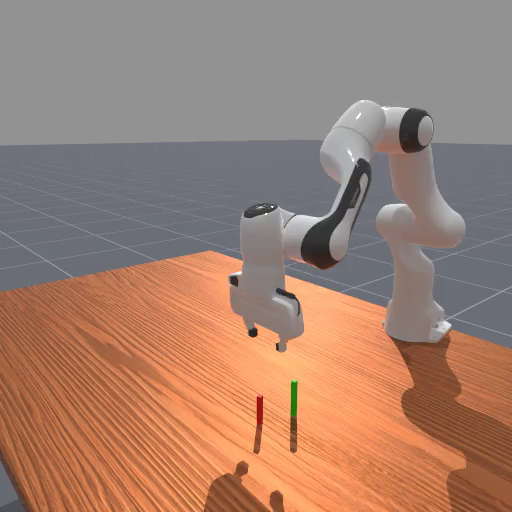}
        \caption{Stack Cuboid}
        \label{fig:task_stack_cuboid}
    \end{subfigure}
    \begin{subfigure}{0.3\textwidth}
        \centering
        \includegraphics[width=0.95\linewidth]{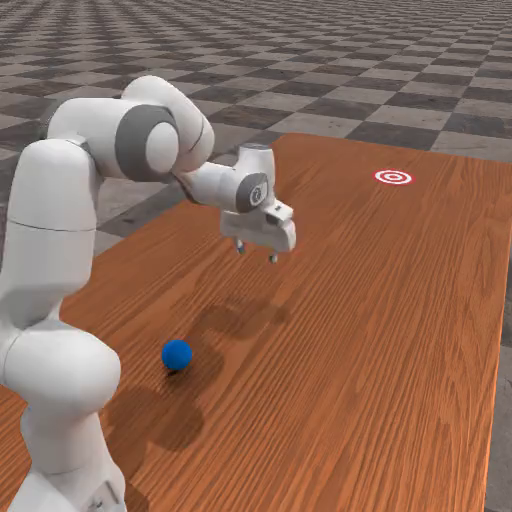}
        \caption{Roll Ball}
        \label{fig:task_roll_ball}
    \end{subfigure}
    
    \caption{\textbf{Illustration of the diverse, high-precision manipulation tasks investigated in this study.} \emph{Peg Insertion:} A contact-rich assembly task requiring precise alignment to overcome tight clearances. \emph{Cuboid Stacking:} A precise placement task where stability is challenged by stacking slender cuboids with a small support base. \emph{Roll Ball:} An underactuated dynamics task requiring control over a rolling ball to reach a distant target.}
    \label{fig:task_description}
\end{figure*}

The pursuit of scaling has become a central theme driving progress in machine learning, leading to powerful foundation models in natural language processing and computer vision \cite{Kaplan2020Scaling,Brown2020Language,Zhai2022ScalingViT,Hestness2017ScalingDL}. The field of robotics is actively seeking to establish its own scaling laws, with the goal of creating more general and capable agents. Recent landmark studies have begun to uncover how performance scales with data diversity, such as the number of objects and environments, aiming to imbue robots with robust zero-shot generalization capabilities in open-world settings \cite{Lin2024Data,Brohan2022RT1,Brohan2023RT2,Padalkar2023OXE}.

However, a vast and economically significant class of robotic applications, ranging from industrial assembly and electronics manufacturing to surgical automation \cite{Siciliano2016Springer}, operates under a different paradigm. In these structured, ``closed-world'' environments, the primary challenge is not generalization to novelty, but the scaling of precision and reliability. For these tasks, improving performance from a 99\% to a 99.9\% success rate, or tightening tolerance from millimeters to sub-millimeters, is of paramount importance. While high precision has traditionally been achieved through analytical control methods \cite{Khatib1987Unified,Hogan1985Impedance}, the quantitative relationship between the data required for imitation learning and the push for higher precision remains largely uncharacterized. The fundamental data cost of achieving precision is still an open question.

Our own initial experiments highlighted this gap: when applying standard imitation learning techniques to high-precision tasks, performance was often unexpectedly low, yet a slight reduction in the task's required precision could lead to a dramatic improvement in success rates. This puzzling sensitivity motivated us to ask: What is the data cost required to increase the precision of a robotic skill? Does a predictable scaling law govern this relationship? And perhaps more importantly, how do the intrinsic capabilities of the learning system itself—its sensors, its expert demonstration quality, and its task complexity—modulate this law?

To answer these questions, we conduct a large-scale empirical study in the ManiSkill3 \cite{Tao2025ManiSkill3} high-fidelity simulation environment, involving over 100 independent model training runs and executing tens of thousands of evaluation rollouts under a rigorous protocol. Our study investigates the three diverse, high-precision manipulation tasks illustrated in Figure~\ref{fig:task_description}. Our primary analysis focuses on two image-based tasks, \emph{Peg Insertion} and \emph{Stack Cuboid}. We then present a state-based \emph{Roll Ball} task, as a case study to validate the generality of our findings. Our extensive investigation reveals the following surprising results and key contributions:

\begin{itemize}
    \item \textbf{The ``Curse of Precision'':} We uncover a novel scaling law demonstrating that data requirements ($N$) grow super-exponentially as the target precision ($P$) increases, accurately modeled by $\log(N) \propto 1/(P-c)$.

    \item \textbf{A New Metric for System Capability:} We identify the law's key parameter, the limit precision $c$, not as a fixed physical constant, but as a quantifiable metric for the performance ceiling of the entire learning system.

    \item \textbf{A Roadmap for Surpassing Precision Limits:} Through systematic ablations, we demonstrate that this performance ceiling $c$ can be dramatically improved by enhancing system components like sensors and expert data quality, offering a clear path to overcoming precision bottlenecks.

\end{itemize}

Our work provides a new theoretical framework for understanding the data cost of precision and, based on it, establishes a practical methodology for developing, debugging, and engineering more capable and data-efficient manipulation systems.

\section{Related Works}

\subsection{Scaling Laws in Robotics and Beyond}

The exploration of scaling laws \cite{Hestness2017ScalingDL} in robotics is heavily influenced by the revolutionary success of this paradigm in foundational domains like natural language processing (NLP) \cite{Kaplan2020Scaling} and computer vision (CV) \cite{Zhai2022ScalingViT}. Seminal works in these fields established that model performance scales as a predictable power-law with increases in dataset size, model parameters, and compute, leading to the emergent generalization capabilities of large-scale models \cite{Kaplan2020Scaling,Brown2020Language}. Inspired by this, the robotics community has embarked on a large-scale effort to discover analogous laws for embodied intelligence. This effort has predominantly focused on scaling for ``Breadth''—improving zero-shot generalization by increasing data diversity across various axes \cite{Brohan2023RT2,Lin2024Data,Zeng2022Socratic,Padalkar2023OXE,Brohan2022RT1,OctoTeam2024Octo}.

A primary axis of scaling has been data diversity in terms of environments and objects. The work of \cite{Lin2024Data} is the most systematic study in this direction. They empirically demonstrated that a policy's generalization to novel environments and objects follows a power law with the number of training variations. Their key finding, diversity is all you need, suggests that increasing the variety of settings is far more critical than simply amassing more demonstrations within a fixed setting.

Another significant axis is scaling data heterogeneity across tasks and robot embodiments\cite{Padalkar2023OXE,Dasari2020RoboNet}. The goal here is to train a single, generalist policy on massive, pooled datasets from disparate sources. The Open X-Embodiment project \cite{Padalkar2023OXE} and its derived models like RT-X and Octo \cite{OctoTeam2024Octo} epitomize this approach. These works have shown that training on vast, multi-embodiment, multi-task data can lead to positive cross-skill and cross-embodiment transfer. Similarly, models like Gato \cite{Reed2022Gato} aim to create a single agent capable of performing hundreds of diverse tasks.

More recently, scaling has also targeted semantic knowledge by leveraging pre-trained Vision-Language-Models (VLMs). Models like RT-2 \cite{Brohan2023RT2} co-fine-tune a VLM on both web-scale data and robot trajectories. This allows the model to transfer semantic understanding from the web to the physical world, enabling it to follow novel, abstract instructions.

Taken together, this impressive body of work forms a cohesive research thrust focused on scaling for ``Breadth''. These two axes of scaling, Breadth and Depth, are largely orthogonal. A system can excel in generalization, yet its fundamental physical skill precision remains constrained by its manipulation data, as candidly noted in works like RT-2 \cite{Brohan2023RT2}. While prior work provides invaluable insights into scaling for Breadth, the scaling behavior for achieving ``Depth''—i.e., \textbf{pushing precision and reliability to their physical limits—remains a critical and open question} that our work directly addresses.

\subsection{High-Precision Robotic Manipulation}

Our work also intersects with the extensive literature on high-precision and contact-rich manipulation. Historically, such tasks were the domain of classical control methods \cite{Murray1994Mathematical,Raibert1981Hybrid}, which established a paradigm where precision was an outcome of analytical correctness, not data. Foundational concepts such as Operational Space Control \cite{Khatib1987Unified} and Impedance Control \cite{Hogan1985Impedance} achieve high precision through rigorous analytical models of the robot and its environment, setting a powerful ``zero-data'' precedent.

When the field transitioned to learning-based approaches like Imitation Learning (IL) and Reinforcement Learning (RL), it inherited this precedent but immediately faced a new, dominant bottleneck: the immense cost, time, and safety risks of real-world data collection \cite{Kober2013Survey}. This practical constraint, combined with the high bar set by classical methods, gave rise to a powerful research imperative: \emph{data efficiency} \cite{Schaal1997Learning,Argall2009SurveyLFD,Inoue2017Assembly}.

This pursuit spurred innovations in two main thrusts. The first is algorithmic, aiming to learn more from limited data \cite{Lynch2019Play,Chelsea2017MetaLearning,Shi2023Waypoint,Shridhar2022PerAct,Jiang2023VIMA,Gubbi2020GAIL,Ankile2024ResiP} via sample-efficient architectures like Transformers \cite{Jiang2023VIMA}, advanced paradigms like adversarial imitation learning \cite{Gubbi2020GAIL}, or modified objectives like residual RL \cite{Ankile2024ResiP}. The second thrust is system-level, generating richer data by using simulation \cite{Tobin2017Domain,James2019Sim,Haugaard2021Fast}, algorithmically expanding datasets with corrective examples \cite{Ankile2024JUICER}, or incorporating human feedback \cite{Luo2024HILSERL}. The primary contribution of these works is often the novel, task-specific algorithm or system architecture itself. This frames the research question as ``can our novel method solve this hard problem?'', solidifying the treatment of precision as a static goalpost. 

The research in this paper adopts a fundamentally different scientific posture. Our objective is not to propose a new data-efficient algorithm to surpass a specific benchmark. Instead, we aim to characterize the universal scaling law that connects data requirements to physical task precision. To achieve this, our methodology is built on a different set of principles. We intentionally fix the learning algorithm to a standard Diffusion Policy implementation. By holding the learning method constant, we can \textbf{isolate and measure the intrinsic data cost imposed by ``precision'' itself} as we treat it as a continuous physical variable. This allows us to uncover the fundamental scaling properties of precision manipulation, which is the precise gap our work aims to fill.

\section{Methodology}

This section details the formalisms, core components, and learning paradigm of our study. We first introduce our scaling law formulations and the metric used for evaluation, then describe the robotic learning system that allows us to empirically investigate these laws.

\subsection{Scaling Law Formulation for Precision}

Our study investigates the relationship between three core dimensions: \emph{Data Volume} ($N$), the number of imitation learning trajectories; \emph{Task Precision} ($P$), the required tolerance for successful task completion; and \emph{System Configuration}, which encompasses the agent's intrinsic properties like sensors and policies.

\textbf{Evaluation Metric.} The primary metric for our study is the \emph{Success Rate (SR)}, a binary measure of task completion. For high-reliability applications such as industrial assembly, where near-perfect performance is often required, SR directly answers the critical question of an agent's deployment-readiness. This choice is motivated by our use of simulation, which permits a large number of automated rollouts to obtain a statistically robust, binary measure of reliability. This contrasts with real-world experiments, where rollouts are expensive and a more granular, human-assigned score can extract richer information from fewer trials \cite{Lin2024Data}.We also considered the Mean Squared Error (MSE) between the policy's actions and the expert's as an alternative metric. However, we found that MSE does not always reliably correlate with the final task success, a finding consistent with recent large-scale manipulation studies \cite{Lin2024Data}.
Therefore, we selected SR as our primary metric due to its direct and unambiguous reflection of a policy's practical reliability.

\textbf{Hypothesis 1: The Success Rate Scaling Law.} Our first hypothesis posits that for a fixed task precision $P$, the performance scales with the data volume $N$. This is inspired by the well-established power-law relationships between performance and data in machine learning. A similar phenomenon was recently observed in robotic manipulation for generalization \cite{Lin2024Data}. We adapt this principle and hypothesize that the failure rate $(1-SR)$ follows a power law with $N$, which is linear in log-log space:

$$\log (1-SR) = a \cdot \log(N) + b , \ (\text{at target} \ P)$$

where $a$ and $b$ are fitting parameters.

\textbf{Hypothesis 2: The Precision Scaling Law.} Our second hypothesis introduces a novel scaling law to describe the data volume $N$ required to achieve a given precision $P$. For a fixed target success rate, we hypothesize this relationship is governed by the proximity of $P$ to a theoretical limit precision $c$:

$$\log(N) = m \cdot \frac{1}{P-c} + n , \ (\text{at target} \ SR)$$

where $m$ and $n$ are fitting parameters specific to this model. The parameter $c$ represents the highest possible precision the system can achieve, which is an intrinsic property of the system configuration, independent of the target success rate. As $P$ approaches this limit, the required data volume is expected to grow super-exponentially.

\subsection{Robotic Learning System}

\textbf{Robot and Environment.} All experiments are conducted in the \emph{ManiSkill3} \cite{Tao2025ManiSkill3} simulation benchmark, which is built upon the \emph{SAPIEN} \cite{Xiang2020SAPIEN} physics engine. The agent is a \emph{Franka Emika Panda} 7-DoF arm equipped with a parallel-jaw gripper.

\textbf{Imitation Learning Paradigm.}  We formulate the task as a behavior cloning problem, where the goal is to learn a policy $\pi(a|o)$ that maps an observation $o$ to an action $a$.

\begin{itemize}
    \item \textbf{Observation Space ($o$):} The policy receives multi-modal input consisting of: (1) two 256x256 RGB-D images from a static third-person camera and a wrist-mounted camera, and (2) the robot's proprioceptive state, including its joint angles and end-effector pose.
    \item \textbf{Action Space ($a$):} The policy outputs a 7-DoF action $a = (\Delta x,\ \Delta y,\ \Delta z,\ \Delta roll,\ \Delta pitch,\ \Delta yaw,\ gripper)$ representing the desired end-effector delta pose.
\end{itemize}

\textbf{Learning Algorithm.} To learn the policy $\pi(a|o)$, we use \emph{Diffusion Policy} \cite{Chi2023DP}. We chose this generative approach over simpler regression-based methods for its strong performance and its ability to model complex, multi-modal action distributions, which is crucial for contact-rich precision tasks. Our implementation utilizes a standard architecture with a \emph{ResNet-18} \cite{He2016Resnet} backbone for visual feature extraction and a \emph{1D U-Net} \cite{Ronneberger2015UNet} as the denoising network, with action inference performed over 100 \emph{DDPM} \cite{Ho2020DDPM} steps.

\textbf{Expert Policy Generation.} We employ two types of expert policies depending on the nature of the task. For structured tasks like \emph{Peg Insertion} and \emph{Stack Cuboid}, expert trajectories are generated using privileged, scripted policies that have access to the ground-truth poses of all objects in the simulation. These scripts define a sequence of key waypoints, and a motion planner is used to generate smooth, collision-free trajectories for the robot's end-effector. Conversely, for a dynamically complex task like \emph{Roll Ball}, where defining an optimal scripted policy is intractable, we instead utilize an expert policy trained via reinforcement learning. This approach also allows us to test the generality of our scaling laws across different expert generation paradigms.

\section{Experiments and Results}

\begin{figure*}[t]
    \centering 
    \textbf{Success Rate Scaling Law}

    \begin{subfigure}[b]{0.32\textwidth}
        \centering
        \includegraphics[width=\linewidth]{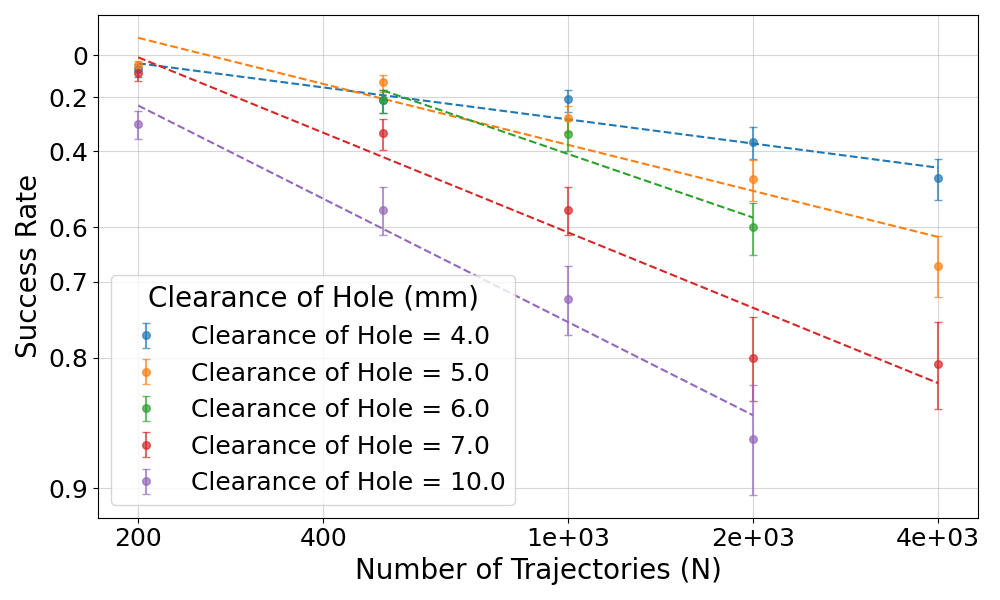}
        \caption{Peg Insertion}
        \label{fig:PegInsertion_fig1} 
    \end{subfigure}
    \hfill 
    \begin{subfigure}[b]{0.32\textwidth}
        \centering
        \includegraphics[width=\linewidth]{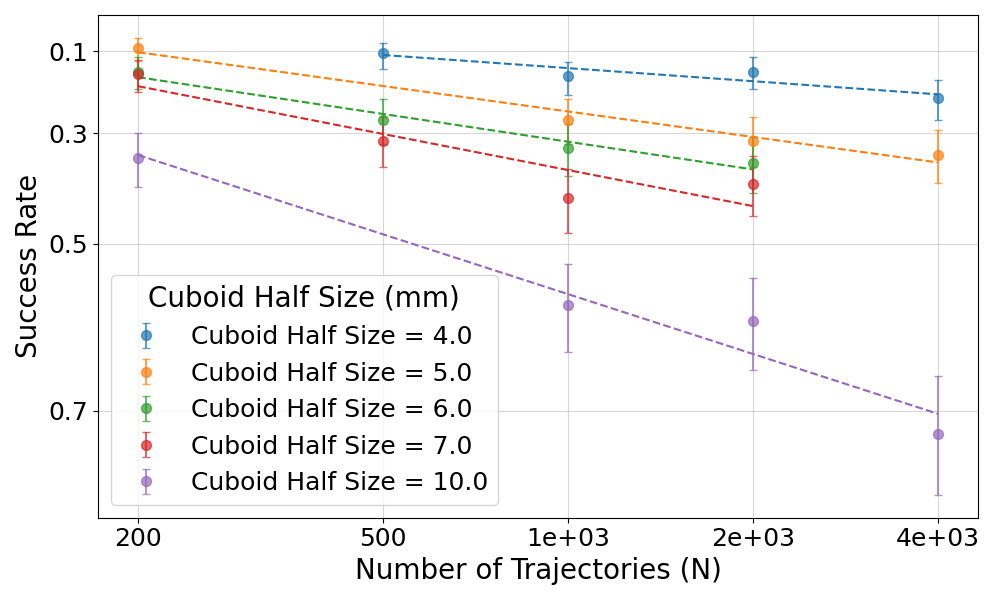}
        \caption{Stack Cuboid}
        \label{fig:StackCuboid_fig1} 
    \end{subfigure}
    \hfill 
    \begin{subfigure}[b]{0.32\textwidth}
        \centering
        \includegraphics[width=\linewidth]{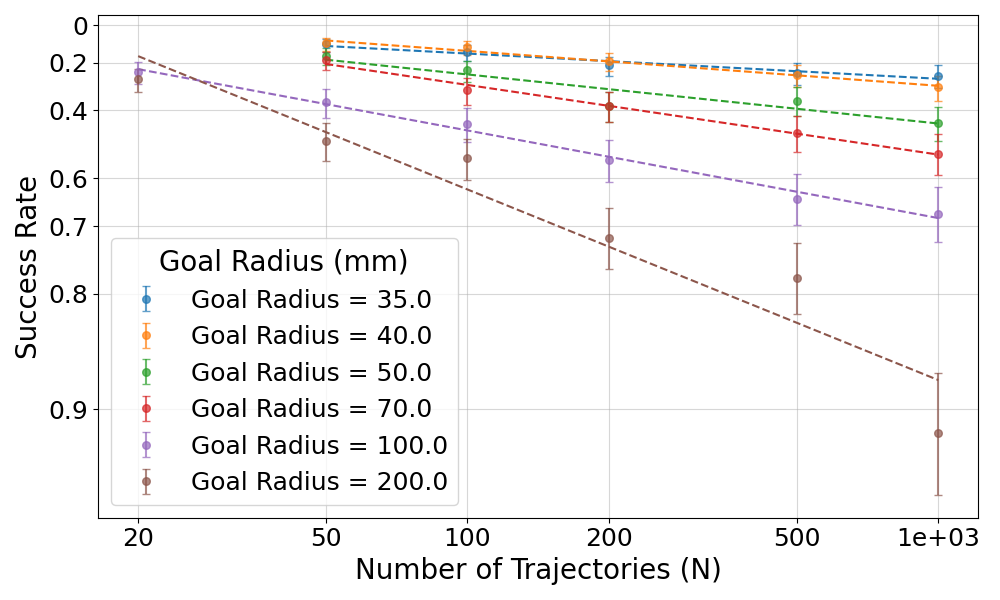}
        \caption{Roll Ball}
        \label{fig:RollBall_fig1} 
    \end{subfigure}

    \vspace{4mm}
    \textbf{Precision Scaling Law}

    \begin{subfigure}[b]{0.32\textwidth}
        \centering
        \includegraphics[width=\linewidth]{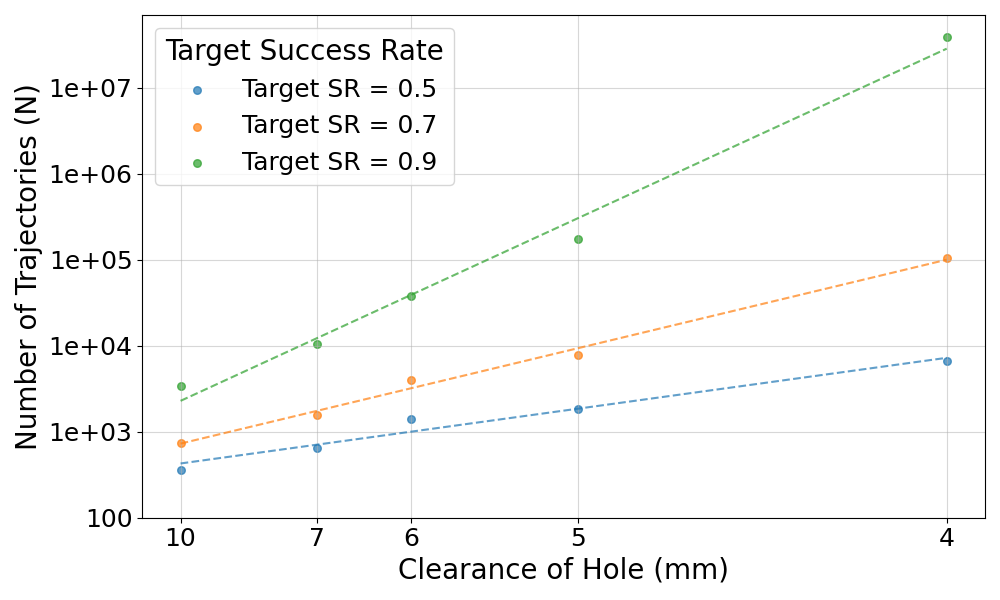}
        \caption{Peg Insertion}
        \label{fig:PegInsertion_fig2} 
    \end{subfigure}
    \hfill 
    \begin{subfigure}[b]{0.32\textwidth}
        \centering
        \includegraphics[width=\linewidth]{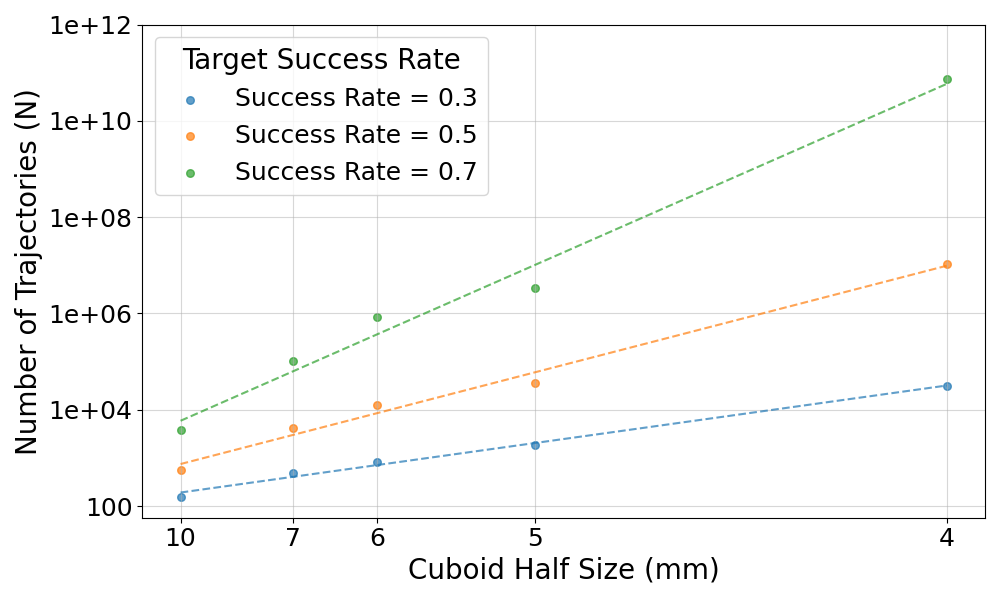}
        \caption{Stack Cuboid}
        \label{fig:StackCuboid_fig2} 
    \end{subfigure}
    \hfill 
    \begin{subfigure}[b]{0.32\textwidth}
        \centering
        \includegraphics[width=\linewidth]{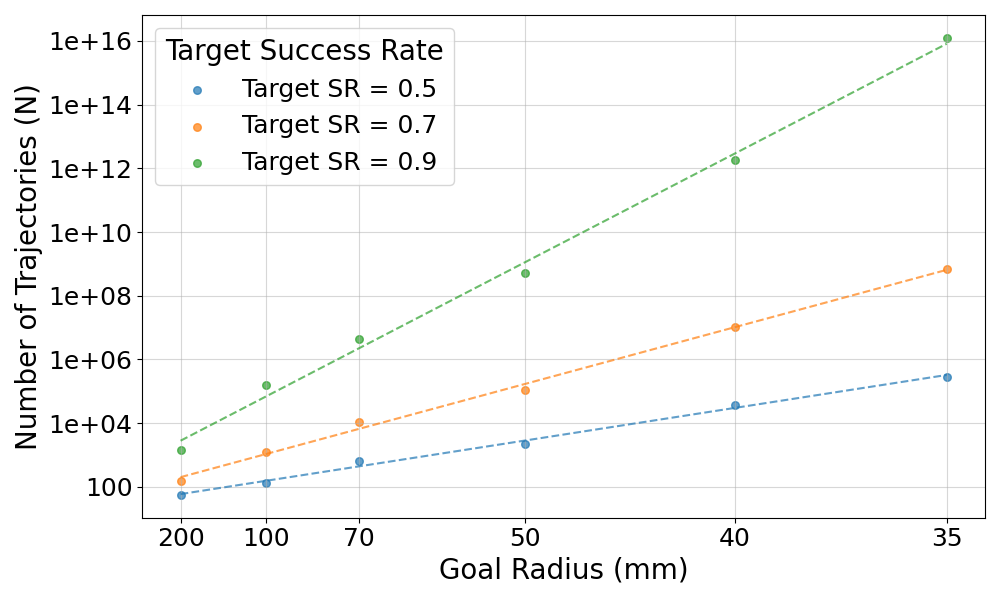}
        \caption{Roll Ball}
        \label{fig:RollBall_fig2} 
    \end{subfigure}
    
    \caption{Empirical Validation of the Proposed Scaling Laws.}
    
    \label{fig:Scaling_Law} 
\end{figure*}

This section presents the empirical validation of our proposed scaling laws. We first detail the experimental tasks and evaluation protocol, then present the results that validate our two hypotheses, and finally investigate how the scaling law parameters are influenced by different system configurations.

\subsection{Tasks and Evaluation Protocol}

\textbf{Tasks.} We validate our hypotheses on three canonical precision manipulation tasks. To ensure the learned skills are robust, both tasks incorporate extensive domain randomization.

\begin{itemize}
    \item \textbf{Peg Insertion:} The task is to insert a square-profile cuboid peg into a corresponding square hole. The precision $P$ is defined as the clearance between the peg and the hole, which we vary from 4mm to 10mm.
    \item \textbf{Cuboid Stacking:} The task is to stack two slender cuboids. The precision $P$ is defined as the half-side-length of the cuboids' square base, which we vary from 4mm to 10mm.
    \item \textbf{Roll Ball:} An underactuated dynamics task where the agent must roll a ball such that it passes through a designated circular target region. The precision $P$ for this task is defined by the radius of this target region, which we vary from 35mm to 200mm.
    
\end{itemize}

\textbf{Evaluation Protocol.} To empirically measure the scaling laws, we followed a rigorous protocol:

\begin{itemize}

    \item \textbf{Data Generation:} For each precision level $P$, we generated a data pool consisting only of successful trajectories.

    \item \textbf{Policy Training Sweep:} For a range of data sizes $N$ (e.g., 200, 500, 1000, 2000), we sampled $N$ trajectories to train a Diffusion Policy model. Due to the high computational cost of training (approx. 20 A100-hours per run), each data point ($N$, $P$) in our plots corresponds to a single training run.

    \item \textbf{Statistical Evaluation and Checkpoint Selection:} Each policy was trained until convergence (typically 160k to 200k steps), with periodic evaluation. To obtain a stable measure of peak performance while mitigating evaluation noise, we report the mean of the top 3 highest SRs achieved during training. Each evaluation consisted of $M=100$ episodes. For our direct measurement plots ($SR$ vs. $N$), we show the 95\% confidence interval of the SR estimate as error bars, calculated using the Wilson score interval method over the $3M = 300$ evaluation trials. This provides a robust measure of the evaluation's statistical reliability.

    \item \textbf{Curve Fitting:} The resulting $(N, SR)$ data points were used to fit our proposed scaling law models. For the Success Rate Scaling Law, we fitted the parameters of the model $\log(1-SR) = a \cdot \log(N)+b$ via linear regression. For the Precision Scaling Law, we first estimated the data size $N$ required to reach different target success rates (e.g., 0.5, 0.7, 0.9) for each $P$ by interpolating or extrapolating from the fitted linear models of the Success Rate Scaling Law. We then performed a grid search for a single $c$ value that maximized the sum of the coefficients of determination ($R^2$) across all three target $SR$ curves. This enforces the hypothesis that $c$ is a constant property of the system, independent of the target success rate.
\end{itemize}

\begin{table}[t]
    \centering
    \caption{Parameters for the Success Rate Scaling Law}
    \label{tab:success_rate_law_fit}
    \begin{tabular}{lcccc}
        \toprule
        \textbf{Task} & \textbf{Precision (mm)} & \textbf{$a$} & \textbf{$b$} & \textbf{$R^2$} \\
        \midrule
        \multirow{5}{*}{Peg Insertion} & 4.0 & -0.19 & 1.35 & 0.921 \\
                                       & 5.0 & -0.35 & 2.83 & 0.900 \\
                                       & 6.0 & -0.49 & 4.10 & 0.933 \\
                                       & 7.0 & -0.58 & 4.40 & 0.939 \\
                                       & 10.0 & -0.72 & 5.08 & 0.967 \\
        \midrule
        \multirow{4}{*}{Stack Cuboid} & 4.0 & -0.06 & 0.35 & 0.840 \\
                                      & 5.0 & -0.11 & 0.70 & 0.977 \\
                                      & 6.0 & -0.12 & 0.67 & 0.970 \\
                                      & 7.0 & -0.16 & 0.92 & 0.843 \\
                                      & 10.0 & -0.26 & 1.41 & 0.958 \\
        \midrule
         \multirow{6}{*}{Roll Ball} & 35.0 & -0.07 & 0.19 & 0.949 \\
                                    & 40.0 & -0.09 & 0.38 & 0.981 \\
                                    & 50.0 & -0.13 & 0.42 & 0.874 \\
                                    & 70.0 & -0.18 & 0.69 & 0.990 \\
                                    & 100.0 & -0.23 & 0.61 & 0.992 \\
                                    & 200.0 & -0.50 & 1.88 & 0.918 \\
        \bottomrule
    \end{tabular}
\end{table}

\begin{table}[t]
    \centering
    \caption{Parameters for the Precision Scaling Law}
    \label{tab:precision_law_fit}
    \begin{tabular}{lccccc}
        \toprule
        \textbf{Task} & \textbf{Target SR} & \textbf{$m$} & \textbf{$n$} & \textbf{$c$ (mm)} & \textbf{$R^2$} \\
        \midrule
        \multirow{3}{*}{Peg Insertion}  & 0.5 & 8.59 & 7.62 & \multirow{3}{*}{2.35} & 0.968 \\
                                        & 0.7 & 14.95 & 7.55 & & 0.993 \\
                                        & 0.9 & 28.62 & 7.42 & & 0.989 \\
        \midrule
        \multirow{3}{*}{Stack Cuboid}   & 0.3 & 11.26 & 5.90 & \multirow{3}{*}{2.75} & 0.993 \\
                                        & 0.5 & 20.81 & 6.47 & & 0.988 \\
                                        & 0.7 & 35.30 & 7.35 & & 0.985 \\
        \midrule
        \multirow{3}{*}{Roll Ball}      & 0.5 & 198.95 & 4.77 & \multirow{3}{*}{20.3} & 0.994 \\
                                        & 0.7 & 346.22 & 5.72 & & 0.997 \\
                                        & 0.9 & 662.94 & 7.76 & & 0.996 \\                    
        
        \bottomrule
    \end{tabular}
\end{table}

\subsection{Validation of the Proposed Scaling Laws}

Our first set of experiments aims to validate the two scaling law hypotheses across all three diverse manipulation tasks. By showing that the laws hold for image-based tasks with scripted experts (Peg Insertion, Stack Cuboid) as well as for a state-based task with an RL-trained expert (Roll Ball), we demonstrate the generality of our findings. The results are presented in Figure~\ref{fig:Scaling_Law}, with detailed fit parameters in Table~\ref{tab:success_rate_law_fit} and ~\ref{tab:precision_law_fit}.

\textbf{Success Rate Scaling Law.} As predicted by Hypothesis 1, our results confirm that performance scales with data size following a power law. As shown in Figure~\ref{fig:Scaling_Law}(a-c), all three tasks exhibit a clear power-law relationship between the failure rate $(1-SR)$ and the data size $N$. The detailed fitting parameters in Table~\ref{tab:success_rate_law_fit} further confirm this with high $R^2$ values across the board, underscoring the generality of this principle.

\textbf{Precision Scaling Law.} The central result of our paper is the validation of the Precision Scaling Law across all three tasks. As shown in Figure~\ref{fig:Scaling_Law}(d-f), when we plot $\log(N)$ against $1/(P-c)$ (with fit parameters detailed in Table~\ref{tab:precision_law_fit}), the data points for all tasks and all target success rates align almost perfectly on straight lines. Crucially, for each task, the data points for all three target SRs (e.g., 0.5, 0.7, 0.9) are fitted with a single, shared value of $c$, yet all yield near-perfect linear fits ($R^2 > 0.97$ as shown in Table II). This strongly \textbf{validates our hypothesis that $c$ is an intrinsic property of the system}, independent of the desired performance level.

\subsection{The Limit Precision $c$ as a System Performance Metric}

\begin{figure*}[t!] 
    \centering

    \begin{minipage}[t]{0.3\linewidth}
        \centering
        \mbox{\textbf{Ablation: Observation}}
        
        \vspace{2mm} 
        
        \begin{subfigure}{0.48\linewidth}
            \includegraphics[width=\textwidth]{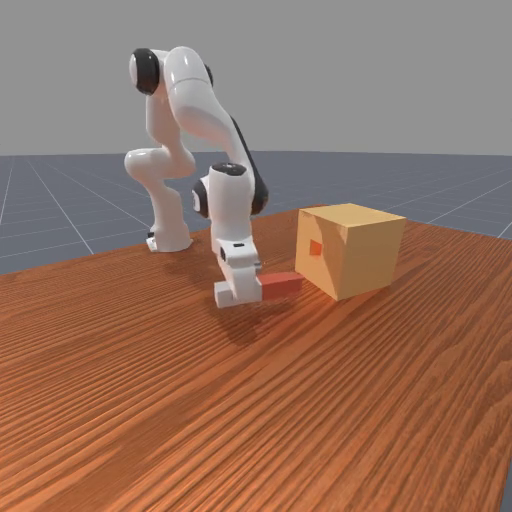}
            \caption{Base Camera}
            \label{fig:ab1_base}
        \end{subfigure}        
        \begin{subfigure}{0.48\linewidth}
            \includegraphics[width=\textwidth]{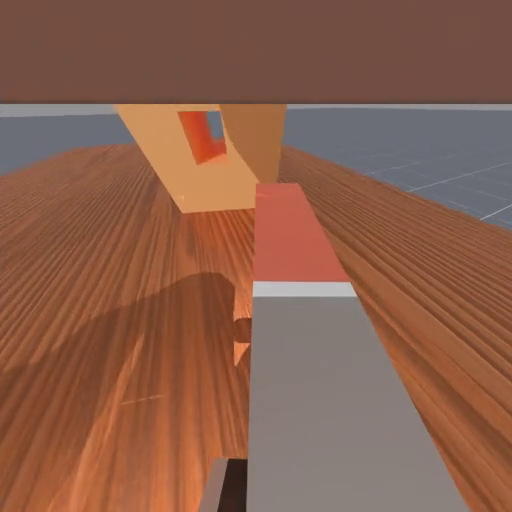}
            \caption{Wrist Camera}
            \label{fig:ab1_hand}
        \end{subfigure}
    \end{minipage}
    \hspace{1em} 
    \begin{minipage}[t]{0.61\linewidth}
        \centering
        \mbox{\textbf{Ablation: Task Complexity}}
        
        \vspace{1.5mm} 
        
        \begin{subfigure}{0.48\linewidth}
            \includegraphics[width=\textwidth]{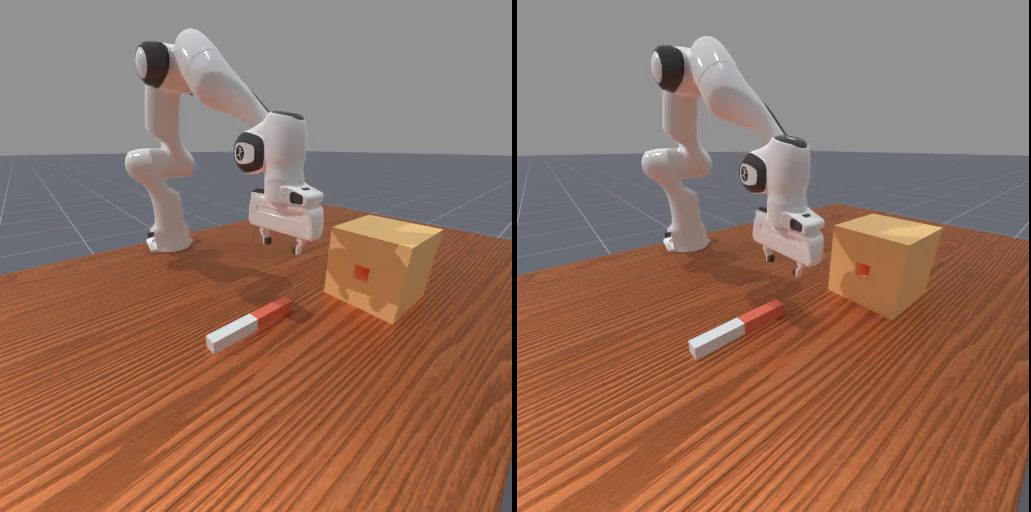}
            \caption{Low Randomization (2 samples)}
            \label{fig:ab3_low}
        \end{subfigure}
        \begin{subfigure}{0.48\linewidth}
            \includegraphics[width=\textwidth]{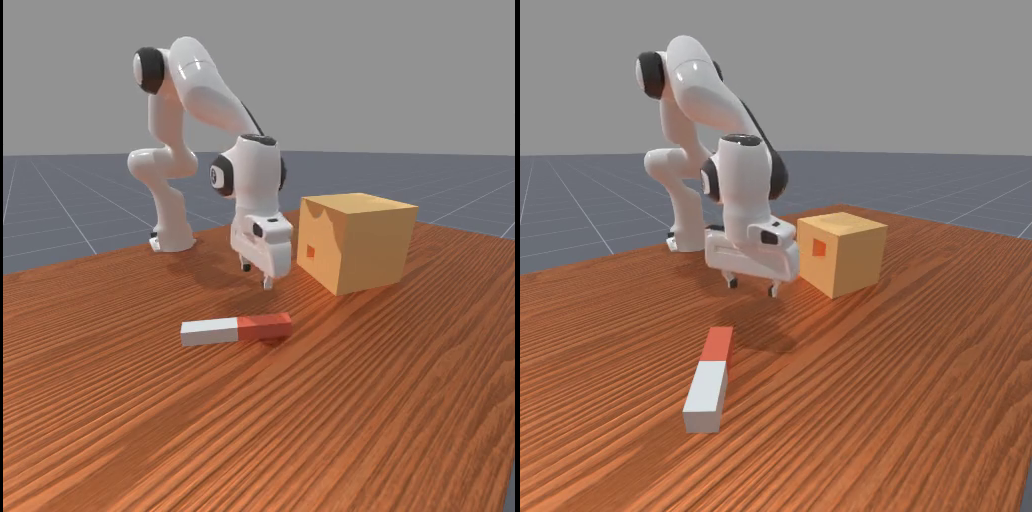}
            \caption{High Randomization (2 samples)}
            \label{fig:ab3_high}
        \end{subfigure}
    \end{minipage}

    \vspace{4mm} 
    
    \begin{minipage}[t]{0.96\linewidth}
        \centering
        \textbf{Ablation: Expert Strategy} \\
        \vspace{1mm}
        
        \begin{subfigure}[b]{0.98\linewidth}
            \centering
            \includegraphics[width=\textwidth]{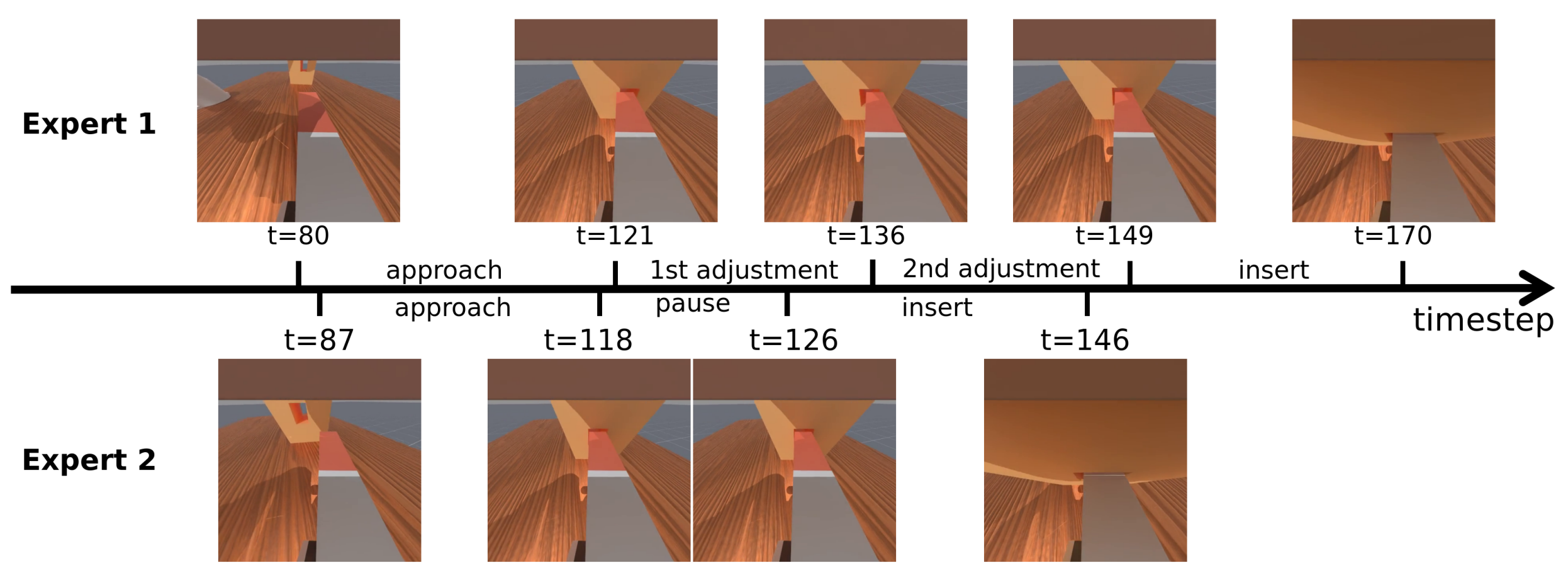}
            \caption{Expert Strategy}
            \label{fig:ablation_expert}
        \end{subfigure}
    \end{minipage}

    \caption{
         \textbf{Visual comparison of the key system configurations investigated in our ablation studies.} (a-b) \textbf{Observation:} The policy's visual input is composed of a (a) base camera and (b) wrist camera. The baseline system utilizes both, while the ablation relies solely on the base camera. (c-d) \textbf{Task Complexity:} We compare two levels of domain randomization. (c) Low Randomization varies only the initial XY positions of the objects. (d) High Randomization, used in our baseline, additionally randomizes object orientations and sizes, hole sizes, and the hole's position on the box face. (e) \textbf{Expert Strategy:} A visual comparison of the two expert strategies. The \emph{Conservative Strategy} (Expert 1, top row) is characterized by several corrective maneuvers prior to insertion. The \emph{Aggressive Strategy} (Expert 2, bottom row) demonstrates a direct strategy, which includes a brief 6-step pause due to the waypoint-based position controller.
    }
    \label{fig:ablation_setup}
\end{figure*}

\begin{table*}[t]
    \centering 
    \begin{threeparttable} 
    
        \caption{The Limit Precision $c$ and Goodness of Fit ($R^2$) under Different System Configurations.}
        \label{tab:system_limit}

        \begin{tabular*}{\textwidth}{@{\extracolsep{\fill}}lcccc}
            \toprule
            \textbf{System Configuration} & \textbf{Limit Precision $c$ (mm)} & \textbf{$R^2$ ($SR=0.5$)} & \textbf{$R^2$ ($SR=0.7$)} & \textbf{$R^2$ ($SR=0.9$)} \\
            \midrule
            \textbf{Baseline} & 2.35 & 0.968 & 0.993 & 0.989 \\
            \textbf{Ablation: Expert Strategy} & 1.27 & 0.986 & 0.988 & 0.979 \\
            \textbf{Ablation: Observation} & 3.85 & 0.983 & 0.997 & 0.987 \\
            \textbf{Ablation: Task Complexity} & 1.00 & 0.996 & 0.992 & 0.981 \\
            
            \bottomrule
        \end{tabular*}

        \begin{tablenotes}
            \small 
            \item[*] The \textbf{Baseline} configuration uses the \emph{Conservative Strategy} (Expert 1) with a wrist camera under high randomization. Each subsequent row represents an ablation study against this baseline. \textbf{Expert Strategy} substitutes Expert 1 with the \emph{Aggressive Strategy} (Expert 2); \textbf{Observation} removes the wrist camera; \textbf{Task Complexity} reduces the level of domain randomization. 
        \end{tablenotes}
        
    \end{threeparttable}
\end{table*}



Having validated the basic form of the scaling laws, we now investigate if the \emph{Precision Scaling Law} holds across different system configurations, and whether its key parameter, the limit precision $c$, acts as a metric for the system's intrinsic capability. To this end, we use the Peg Insertion task as a case study and conduct a series of ablation studies, summarized in Table~\ref{tab:system_limit}. The results provide strong evidence for both points: the law remains an excellent fit ($R^2 > 0.97$) across all configurations, and the consistent value of $c$ found for each setup validates its role as a system-specific metric, allowing us to analyze how it is influenced by different system components.

\textbf{Impact of Observation Modality.} We first analyze the impact of observation modality by disabling the wrist-mounted camer (Figure~\ref{fig:ab1_base}, \ref{fig:ab1_hand}). As shown in Table~\ref{tab:system_limit} (Ablation: Observation), removing this crucial local sensor dramatically degrades system capability, increasing the   limit precision $c$ from 2.35 mm to 3.85 mm. This quantifies the critical value of high-quality local observations for precision tasks.

\textbf{Impact of Expert Policy Strategy.} Our baseline system is trained on data from a \emph{conservative expert strategy} (Expert 1). This cautious approach, which performs two additional, corrective alignment maneuvers at the hole's entrance (Figure~\ref{fig:ablation_expert}), ensures a high raw success rate (about 98\% at 5mm clearance) but creates observational ambiguity for the learned policy. While incorporating historical context is a valid and powerful approach to resolve such ambiguities, akin to how large language models leverage longer context windows, our findings reveal a complementary and highly data-efficient principle: that directly reducing the observational ambiguity within the expert demonstrations is a primary lever for improving scaling behavior.

This motivated us to test this principle by designing an \emph{aggressive expert strategy} (Expert 2), which performs a more direct, single-shot insertion (Figure~\ref{fig:ablation_expert}). This design choice represents a trade-off: to obtain observationally unambiguous demonstrations, we forgo the alignment phase, which lowers the expert's own raw success rate (about 50\% at 5mm clearance). However, by filtering for its successful trajectories, we create a dataset of clean demonstrations. 

As shown in Table~\ref{tab:system_limit} (Ablation: Expert Strategy), the impact is profound: training on the cleaner data from the aggressive expert results in a dramatically improved limit precision c of 1.27 mm. This key finding demonstrates that for imitation learning, \textbf{the clarity and lack of ambiguity in demonstrations can be a more critical factor for achieving high precision} than the expert's own raw success rate.

\textbf{Impact of Task Complexity.} Finally, we investigate task complexity by reducing the level of domain randomization (Fig.~\ref{fig:ab3_low}, \ref{fig:ab3_high}). We only randomize object positions, ensuring the agent must still learn a closed-loop policy. As expected, this simplification is reflected in a dramatically improved limit precision $c$ of 1.00 mm, indicating greater data efficiency in a smaller problem space.

These results collectively provide strong evidence that the limit precision $c$ is a sensitive and informative metric for the overall capability of a robotic learning system.

\subsection{Further Analysis: On the Precondition of Model Capacity}
\label{sec:model_capacity}

While validating the scaling laws, the \emph{Roll Ball} task revealed a crucial insight into a precondition for these laws to emerge: the policy network must have sufficient representational capacity.

\begin{figure}[t]
    \centering
    \includegraphics[width=\columnwidth]{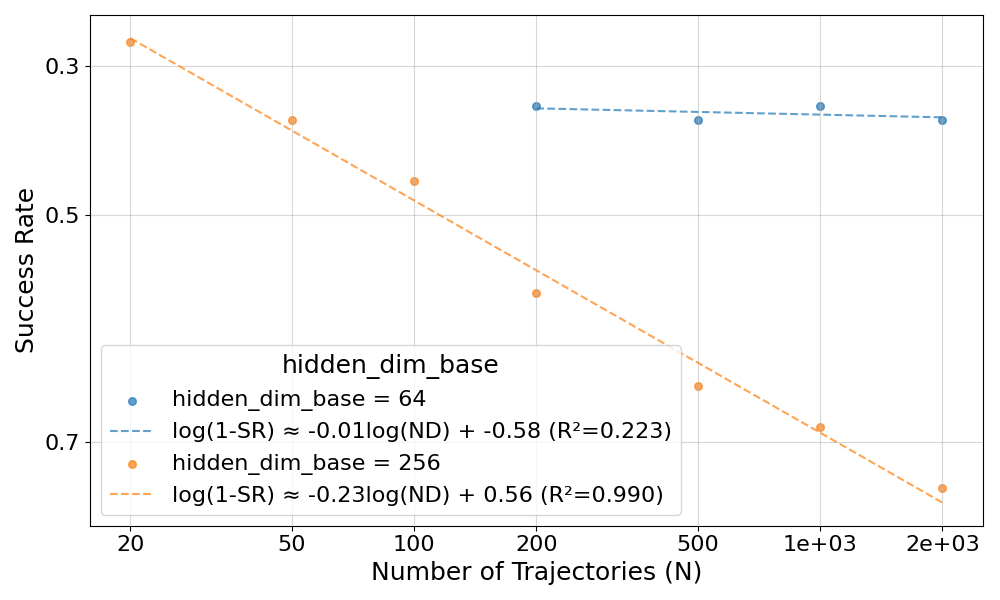}
    \caption{Impact of model capacity on the Success Rate Scaling Law for the Roll Ball task, comparing a policy trained with a   standard-capacity (hidden\_dim\_base=64) versus a high-capacity (hidden\_dim\_base=256) denoiser.}
    \label{fig:capacity_comparison}
\end{figure}

This phenomenon is illustrated in Figure~\ref{fig:capacity_comparison}. In our initial experiments using a standard-capacity 1D U-Net denoiser (blue line), a clear scaling law failed to emerge, with a corresponding coefficient of determination of only $R^2=0.22$. The policy's success rate remained low and did not improve predictably with more data. It was only after we substantially increased the capacity of the denoiser network that a distinct scaling law became apparent (orange line), yielding an almost perfect fit with $R^2=0.99$.

This finding suggests that for certain dynamically complex tasks, sufficient model capacity is a prerequisite for the scaling law to manifest at all. The system's representational power must first be adequate to capture the fundamental dynamics of the task before the predictable relationship between data and precision can take hold.

\subsection{Discussion}

The empirical findings from our scaling law validation and ablation studies carry significant implications for practical robotic engineering. Our work primarily reveals that for high-precision tasks, ``data is not all you need''. When approaching a system's intrinsic precision limit $c$, simply scaling up the data volume yields diminishing returns due to the ``Curse of Precision''.

More importantly, our scaling law provides a powerful \emph{predictive tool} and a \emph{debugging tool for high-precision tasks}. A key distinction must be made between the law's one-time, data-intensive \emph{validation} (the focus of this paper) and its far more efficient \emph{application} as a diagnostic. Once the law's form is assumed, engineers can fit its parameters with relatively few experiments to: 1) Predict the ultimate precision limit ($c$) of their current system, and 2) Estimate the data cost ($N$) required to reach a target. This predictive capability is crucial for debugging. When a task fails at high precision, engineers can evaluate on lower-precision versions to check if performance degrades consistently with the scaling law. A predictable trend suggests the system is hitting its intrinsic limit $c$, while erratic performance would indicate a bug in the implementation, thus guiding the engineering focus.

\section{Conclusion}

In this work, we shifted the focus of scaling law research in robotics from the open-world paradigm of generalization to the closed-world challenge of precision and reliability. We investigated the fundamental data cost of achieving high precision in manipulation, a critical but underexplored area. Our large-scale empirical study answered the key questions posed in our introduction: we discovered a predictable, super-exponential scaling law governing the relationship between data and precision. We demonstrated that the intrinsic capabilities of the learning system modulate this law through a core parameter, the limit precision $c$. Finally, we showed that this limit is not a fixed barrier but a quantifiable metric that can be improved through targeted enhancements to system components like sensors and expert data quality.

Our primary contribution is a new theoretical framework and a practical diagnostic tool to guide future efforts away from brute-force data collection and towards more intelligent, system-level solutions. The main takeaway is that for high-precision robotics, the path forward lies not just in scaling data, but in systematically understanding and improving the fundamental limits of our learning systems.

\textbf{Limitations and Future Work.} Our work opens several avenues for future research. First, our ablation study on expert strategies revealed that demonstration clarity, rather than the expert's raw success rate, is a critical factor for achieving high precision. This suggests a promising research direction in algorithmic data curation. Instead of relying on perfect, unambiguous teleoperation, future work could explore methods to automatically ``clean'' or ``edit'' large datasets of imperfect but plentiful demonstrations to extract a smaller set of high-quality, unambiguous trajectories. This could provide a more scalable path to improving a system's precision limit $c$. Second, all experiments were conducted in simulation; validating these laws on physical hardware is a crucial next step. Third, our study focuses exclusively on Behavior Cloning. Investigating how these laws change with online methods like Reinforcement Learning or DAgger is another promising direction. Finally, we encourage future work to expand the scope of tasks and system configurations to further validate the universality of the ``Curse of Precision''.

\bibliographystyle{IEEEtran}
\bibliography{references}

@article{Kaplan2020Scaling,
  title={Scaling Laws for Neural Language Models},
  author={Jared Kaplan and Sam McCandlish and Tom Henighan and Tom B. Brown and Benjamin Chess and Rewon Child and Scott Gray and Alec Radford and Jeff Wu and Dario Amodei},
  year={2020},
  journal={arXiv preprint arXiv:2001.08361}
}

@article{Brown2020Language,
  title={Language Models are Few-Shot Learners},
  author={Tom B. Brown and Benjamin Mann and Nick Ryder and Melanie Subbiah and Jared Kaplan and Prafulla Dhariwal and Arvind Neelakantan and Pranav Shyam and Girish Sastry and Amanda Askell and Sandhini Agarwal and Ariel Herbert-Voss and Gretchen Krueger and Tom Henighan and Rewon Child and Aditya Ramesh and Daniel M. Ziegler and Jeffrey Wu and Clemens Winter and Christopher Hesse and Mark Chen and Eric Sigler and Mateusz Litwin and Scott Gray and Benjamin Chess and Jack Clark and Christopher Berner and Sam McCandlish and Alec Radford and Ilya Sutskever and Dario Amodei},
  year={2020},
  journal={arXiv preprint arXiv:2005.14165}
}

@article{Lin2024Data,
  title={Data Scaling Laws in Imitation Learning for Robotic Manipulation},
  author={Fanqi Lin and Yingdong Hu and Pingyue Sheng and Chuan Wen and Jiacheng You and Yang Gao},
  year={2024},
  journal={arXiv preprint arXiv:2410.18647}
}

@article{Brohan2022RT1,
  title={RT-1: Robotics Transformer for Real-World Control at Scale},
  author={Anthony Brohan and Noah Brown and Justice Carbajal and Yevgen Chebotar and Joseph Dabis and Chelsea Finn and Keerthana Gopalakrishnan and Karol Hausman and Alex Herzog and Jasmine Hsu and Julian Ibarz and Brian Ichter and Alex Irpan and Tomas Jackson and Sally Jesmonth and Nikhil Joshi and Ryan Julian and Dmitry Kalashnikov and Yuheng Kuang and Isabel Leal and Kuang-Huei Lee and Sergey Levine and Yao Lu and Utsav Malla and Deeksha Manjunath and and Igor Mordatch and Ofir Nachum and Carolina Parada and Jodilyn Peralta and Emily Perez and Karl Pertsch and Jornell Quiambao and Kanishka Rao and Michael Ryoo and Grecia Salazar and Pannag Sanketi and Kevin Sayed and Jaspiar Singh and Sumedh Sontakke and Austin Stone and Clayton Tan and Huong Tran and Vincent Vanhoucke and Steve Vega and Quan Vuong and Fei Xia and Ted Xiao and Peng Xu and Sichun Xu and Tianhe Yu and Brianna Zitkovich},
  year={2022},
  journal={arXiv preprint arXiv:2212.06817}
}

@article{Padalkar2023OXE,
  title={Open X-Embodiment: Robotic Learning Datasets and RT-X Models},
  author={Abhishek Padalkar and others},
  year={2023},
  journal={arXiv preprint arXiv:2310.08864}
}

@book{Siciliano2016Springer,
  title={Springer Handbook of Robotics},
  editor={Bruno Siciliano and Oussama Khatib},
  year={2016},
  publisher={Springer},
  edition={2nd}
}

@article{OctoTeam2024Octo,
  title={Octo: An Open-Source Generalist Robot Policy},
  author={{Octo Model Team} and Dibya Ghosh and Homer Walke and Karl Pertsch and Kevin Black and Oier Mees and Sudeep Dasari and Joey Hejna and Charles Xu and Jianlan Luo and Tobias Kreiman and {You Liang} Tan and Pannag Sanketi and Quan Vuong and Ted Xiao and Dorsa Sadigh and Chelsea Finn and Sergey Levine},
  year={2024},
  journal={arXiv preprint arXiv:2405.12213}
}

@article{Reed2022Gato,
  title={A Generalist Agent},
  author={Scott Reed and Konrad Zolna and Emilio Parisotto and Sergio Gomez Colmenarejo and Alexander Novikov and Gabriel Barth-Maron and Mai Gimenez and Yury Sulsky and Jackie Kay and Jost Tobias Springenberg and Tom Eccles and Jake Bruce and Ali Razavi and Ashley Edwards and Nicolas Heess and Yutian Chen and Raia Hadsell and Oriol Vinyals and Mahyar Bordbar and Nando de Freitas},
  year={2022},
  journal={arXiv preprint arXiv:2205.06175}
}

@inproceedings{Xiang2020SAPIEN,
  title={{SAPIEN}: A SimulAted Part-based Interactive ENvironment},
  author={Fanbo Xiang and Yuzhe Qin and Kaichun Mo and Yikuan Xia and Hao Zhu and Fangchen Liu and Minghua Liu and Hanxiao Jiang and Yifu Yuan and He Wang and Li Yi and Angel Chang and Leonidas J. Guibas and Hao Su},
  booktitle={Proceedings of the IEEE/CVF Conference on Computer Vision and Pattern Recognition (CVPR)},
  year={2020}
}

@inproceedings{Chi2023DP,
  author = {Cheng Chi and Siyuan Feng and Yilun Du and Zhenjia Xu and Eric Cousineau and Benjamin Burchfiel and Shuran Song},
  title = {{Diffusion Policy: Visuomotor Policy Learning via Action Diffusion}},
  booktitle = {Proceedings of Robotics: Science and Systems},
  year = {2023},
  address = {Daegu, Republic of Korea},
  month = {July},
  doi = {10.15607/RSS.2023.XIX.026}
}

@inproceedings{He2016Resnet,
  title={Deep Residual Learning for Image Recognition},
  author={Kaiming He and Xiangyu Zhang and Shaoqing Ren and Jian Sun},
  booktitle={Proceedings of the IEEE conference on computer vision and pattern recognition},
  year={2016}
}

@article{Tao2025ManiSkill3,
  title={ManiSkill3: GPU Parallelized Robotics Simulation and Rendering for Generalizable Embodied AI},
  author={Stone Tao and Fanbo Xiang and Arth Shukla and Yuzhe Qin and Xander Hinrichsen and Xiaodi Yuan and Chen Bao and Xinsong Lin and Yulin Liu and Tse-kai Chan and Yuan Gao and Xuanlin Li and Tongzhou Mu and Nan Xiao and Arnav Gurha and Viswesh Nagaswamy Rajesh and Yong Woo Choi and Yen-Ru Chen and Zhiao Huang and Roberto Calandra and Rui Chen and Shan Luo and Hao Su},
  journal = {Robotics: Science and Systems},
  year={2025}
}

@article{Khatib1987Unified,
  author = {Oussama Khatib},
  title = {A Unified Approach for Motion and Force Control of Robot Manipulators: The Operational Space Formulation},
  journal = {IEEE Journal on Robotics and Automation},
  year = {1987},
  volume = {3},
  number = {1},
  pages = {43-53},
  doi = {10.1109/JRA.1987.1087068}
}

@article{Hogan1985Impedance,
  author = {Neville Hogan},
  title = {Impedance Control: An Approach to Manipulation, Part I: Theory},
  journal = {Journal of Dynamic Systems, Measurement, and Control},
  year = {1985},
  volume = {107},
  number = {1},
  pages = {1-7},
  doi = {10.1115/1.3140702}
}

@article{Kober2013Survey,
  author = {Jens Kober and J. Andrew Bagnell and Jan Peters},
  title = {Reinforcement Learning in Robotics: A Survey},
  journal = {The International Journal of Robotics Research},
  year = {2013},
  volume = {32},
  number = {11},
  pages = {1238-1274},
  doi = {10.1177/0278364913495721}
}

@article{Ankile2024ResiP,
  author = {Lars Lien Ankile and Anthony Simeonov and Idan Shenfeld and Marcel Torne Villasevil and Pulkit Agrawal},
  title = {From Imitation to Refinement -- Residual RL for Precise Visual Assembly},
  journal = {arXiv preprint arXiv:2407.16677},
  year = {2024}
}

@inproceedings{Tobin2017Domain,
  title={Domain randomization for transferring deep neural networks from simulation to the real world},
  author={Joshua Tobin and Rachel Fong and Alex Ray and Jonas Schneider and Wojciech Zaremba and Pieter Abbeel},
  booktitle={2017 IEEE/RSJ International Conference on Intelligent Robots and Systems (IROS)},
  pages={23--30},
  year={2017},
  organization={IEEE}
}

@article{Ankile2024JUICER,
  author = {Lars Ankile and Anthony Simeonov and Idan Shenfeld and Pulkit Agrawal},
  title = {JUICER: Data-Efficient Imitation Learning for Robotic Assembly},
  journal = {arXiv preprint arXiv:2402.14833},
  year = {2024}
}

@article{Luo2024HILSERL,
  author = {Jianlan Luo and Charles Xu and Jeffrey Wu and Sergey Levine},
  title = {Precise and Dexterous Robotic Manipulation via Human-in-the-Loop Reinforcement Learning},
  journal = {arXiv preprint arXiv:2410.21845},
  year = {2024}
}

@inproceedings{Jiang2023VIMA,
  title={VIMA: General Robot Manipulation with Multimodal Prompts},
  author={Yunfan Jiang and Agrim Gupta and Zichen Zhang and Guanzhi Wang and Yongqiang Dou and Yanjun Chen and Li Fei-Fei and Anima Anandkumar and Yuke Zhu and Linxi Fan},
  booktitle={International Conference on Machine Learning},
  year={2023}
}

@inproceedings{Gubbi2020GAIL,
  author={S. Gubbi and Shishir N. Y. Kolathaya and B. Amrutur},
  booktitle={2020 6th International Conference on Control, Automation and Robotics (ICCAR)},
  title={Imitation Learning for High Precision Peg-in-Hole Tasks},
  year={2020},
  pages={1-6},
  doi={10.1109/ICCAR49649.2020.9096700}
}

@inproceedings{Ronneberger2015UNet,
  author    = {Olaf Ronneberger and Philipp Fischer and Thomas Brox},
  title     = {U-Net: Convolutional Networks for Biomedical Image Segmentation},
  booktitle = {Medical Image Computing and Computer-Assisted Intervention -- MICCAI 2015},
  pages     = {234--241},
  year      = {2015},
  publisher = {Springer International Publishing},
  series    = {Lecture Notes in Computer Science},
  volume    = {9351},
  doi       = {10.1007/978-3-319-24574-4_28}
}

@article{Ho2020DDPM,
  title={Denoising Diffusion Probabilistic Models},
  author={Jonathan Ho and Ajay Jain and Pieter Abbeel},
  year={2020},
  journal={arXiv preprint arXiv:2006.11239}
}

@article{Hestness2017ScalingDL,
  title={Deep learning scaling is predictable, empirically},
  author={Hestness, Joel and Narang, Sharan and Ardalani, Newsha and Diamos, Gregory and Jun, Heewoo and Kianinejad, Hassan and Patwary, Mostofa Ali and Yang, Yang and Zhou, Yanqi},
  journal={arXiv preprint arXiv:1712.00409},
  year={2017}
}

@InProceedings{Zhai2022ScalingViT,
    author = {Zhai, Xiaohua and Kolesnikov, Alexander and Houlsby, Neil and Beyer, Lucas},
    title = {Scaling Vision Transformers},
    booktitle = {Proceedings of the IEEE/CVF Conference on Computer Vision and Pattern Recognition (CVPR)},
    month = {June},
    year = {2022},
    pages = {12104-12113}
}

@article{Brohan2023RT2,
  author = {Brohan, Anthony and Brown, Noah and Carbajal, Justice and Chebotar, Yevgen and Chen, Xi and Choromanski, Krzysztof and Ding, Tianli and Driess, Danny and Dubey, Avinava and Finn, Chelsea and Florence, Pete and others},
  title = {RT-2: Vision-Language-Action Models Transfer Web Knowledge to Robotic Control},
  journal = {arXiv preprint arXiv:2307.15818},
  year = {2023}
}

@article{Zeng2022Socratic,
  title={Socratic Models: Composing Zero-Shot Multimodal Reasoning with Language},
  author={Zeng, Andy and Attarian, Maria and Ichter, Brian and Choromanski, Krzysztof and Wong, Adrian and Welker, Stefan and Tombari, Federico and Purohit, Aveek and Ryoo, Michael S. and Sindhwani, Vikas and Lee, Johnny and Vanhoucke, Vincent and Florence, Pete},
  journal={arXiv preprint arXiv:2204.00598},
  year={2022}
}

@book{Murray1994Mathematical,
  title={A mathematical introduction to robotic manipulation},
  author={Murray, Richard M. and Li, Zexiang and Sastry, S. Shankar},
  year={1994},
  publisher={CRC press}
}

@article{Raibert1981Hybrid,
  title={Hybrid position/force control of manipulators},
  author={Raibert, Marc H. and Craig, John J.},
  journal={Journal of dynamic systems, measurement, and control},
  volume={103},
  number={2},
  pages={126--133},
  year={1981},
  publisher={American Society of Mechanical Engineers}
}

@inproceedings{Schaal1997Learning,
  title = {Learning from demonstration},
  booktitle = {Advances in Neural Information Processing Systems 9},
  author = {Schaal, S.},
  pages = {1040--1046},
  year = {1997},
  publisher = {MIT Press},
  address = {Cambridge, MA}
}

@article{Argall2009SurveyLFD,
  title={A survey of robot learning from demonstration},
  author={Argall, Brenna D and Chernova, Sonia and Veloso, Manuela and Browning, Brett},
  journal={Robotics and autonomous systems},
  volume={57},
  number={5},
  pages={469--483},
  year={2009},
  publisher={Elsevier}
}

@InProceedings{Chelsea2017MetaLearning,
  title = {Model-Agnostic Meta-Learning for Fast Adaptation of Deep Networks},
  author = {Chelsea Finn and Pieter Abbeel and Sergey Levine},
  booktitle = {Proceedings of the 34th International Conference on Machine Learning},
  pages = {1126--1135},
  year = {2017},
  editor = {Precup, Doina and Teh, Yee Whye},
  volume = {70},
  series = {Proceedings of Machine Learning Research},
  month = {06--11 Aug},
  publisher = {PMLR}
}

@inproceedings{Lynch2019Play,
  title = {Learning Latent Plans from Play},
  author = {Lynch, Corey and Khansari, Mohi and Xiao, Ted and Kumar, Vikash and Tompson, Jonathan and Levine, Sergey and Sermanet, Pierre},
  booktitle = {Proceedings of the Conference on Robot Learning (CoRL)},
  year = {2019}
}

@InProceedings{James2019Sim,
author = {James, Stephen and Wohlhart, Paul and Kalakrishnan, Mrinal and Kalashnikov, Dmitry and Irpan, Alex and Ibarz, Julian and Levine, Sergey and Hadsell, Raia and Bousmalis, Konstantinos},
title = {Sim-To-Real via Sim-To-Sim: Data-Efficient Robotic Grasping via Randomized-To-Canonical Adaptation Networks},
booktitle = {The IEEE Conference on Computer Vision and Pattern Recognition (CVPR)},
month = {June},
year = {2019}
}

@InProceedings{Dasari2020RoboNet,
  title = {RoboNet: Large-Scale Multi-Robot Learning},
  author = {Dasari, Sudeep and Ebert, Frederik and Tian, Stephen and Nair, Suraj and Bucher, Bernadette and Schmeckpeper, Karl and Singh, Siddharth and Levine, Sergey and Finn, Chelsea},
  booktitle = {Proceedings of the Conference on Robot Learning},
  pages = {885--897},
  year = {2020},
  editor = {Kaelbling, Leslie Pack and Kragic, Danica and Sugiura, Komei},
  volume = {100},
  series = {Proceedings of Machine Learning Research},
  month = {30 Oct--01 Nov},
  publisher = {PMLR}
}

@inproceedings{Inoue2017Assembly,
  title={Deep Reinforcement Learning for High Precision Assembly Tasks},
  author={Inoue, Tadanobu and De Magistris, Giovanni and Munawar, Asim and Yokoya, Tsuyoshi and Tachibana, Ryuki},
  booktitle={2017 IEEE/RSJ International Conference on Intelligent Robots and Systems (IROS)},
  pages={819--825},
  year={2017},
  organization={IEEE},
  doi={10.1109/IROS.2017.8196039},
  arxiv={1708.04033}
}

@article{Shi2023Waypoint,
  title={Waypoint-Based Imitation Learning for Robotic Manipulation},
  author={Lucy Xiaoyang Shi and Archit Sharma and Tony Z. Zhao and Chelsea Finn},
  journal={7th Conference on Robot Learning (CoRL 2023)},
  year={2023}
}

@inproceedings{Shridhar2022PerAct,
  title={Perceiver-Actor: A Multi-Task Transformer for Robotic Manipulation},
  author={Mohit Shridhar and Lucas Manuelli and Dieter Fox},
  booktitle={CoRL},
  year={2022}
}

@inproceedings{Haugaard2021Fast,
  title={Fast robust peg-in-hole insertion with continuous visual servoing},
  author={Haugaard, Rasmus and Langaa, Jeppe and Sloth, Christoffer and Buch, Anders},
  booktitle={Proceedings of the 2020 Conference on Robot Learning},
  pages={1696--1705},
  year={2021},
  organization={PMLR},
  volume={155},
  publisher={PMLR}
}

\end{document}